%% file: main.tex
\documentclass{article}

\usepackage[final]{corl_2020_arxiv} 

\usepackage{graphicx}

\usepackage{times}
\usepackage{epsfig}
\usepackage{amsmath}
\usepackage{amssymb}
\usepackage{arydshln}
\input{packages}
\input{macro}

\title{PLG-IN: Pluggable Geometric Consistency Loss with Wasserstein Distance in Monocular Depth Estimation}
\author{
  Noriaki Hirose,~~~~Satoshi Koide,~~~~Keisuke Kawano,~~~~Ruho Kondo\thanks{The authors are sorted in descending order based on their contributions.}\\
  TOYOTA Central R$\&$D Labs., INC.
  Japan\\
  \texttt{hirose@mosk.tytlabs.co.jp} \\
}

\begin{document}
\maketitle

\begin{abstract}
   We propose a novel objective for penalizing geometric inconsistencies to improve the depth and pose estimation performance of monocular camera images. Our objective is designed using the Wasserstein distance between two point clouds, estimated from images with different camera poses. The Wasserstein distance can impose a soft and symmetric coupling between two point clouds, which suitably maintains geometric constraints and results in a differentiable objective. By adding our objective to the those of other state-of-the-art methods, we can effectively penalize geometric inconsistencies and obtain highly accurate depth and pose estimations. Our proposed method is evaluated using the KITTI dataset.
\end{abstract}


\section{Introduction}
Understanding the three-dimensional (3D) structures of environments and objects is important for the navigation of autonomous vehicles and for robotic manipulation\cite{thrun2002probabilistic,biswas2012depth}. 
In recent years, depth estimation using RGB monocular images has been actively researched owing to the popularity of deep learning. Cameras can function as inexpensive and affordable sensors for autonomous vehicles and robots~\cite{hirose2019deep,hirose2018gonet}.
Hence, they can be viable alternatives to expensive LIDARs. However, it is difficult to obtain a considerable number of depth image ground truths from LIDAR data, for application to machine learning techniques, and data collection is itself a challenge.

Self-supervised learning using videos captured by robots and vehicles enables the learning of depth and pose estimation networks, without ground truth depth images and poses. Here, pose estimation refers to the relative camera pose estimation between two consecutive images. The image reconstruction loss proposed by Zhou et al.~\cite{zhou2017unsupervised} has significantly improved accuracy. Thereafter, various pluggable objectives, network structures, data augmentation methods, and masking methods for dynamic and occluded objects have been suggested to enable improvement~\cite{zhou2017unsupervised,yang2018every,yang2018lego,mahjourian2018unsupervised,wang2018learning,casser2019depth,gordon2019depth,godard2019digging,fei2019geo,pillai2019superdepth}.

In this paper, we propose a novel pluggable objective to evaluate geometric inconsistencies with respect to these prior methods.
As the geometry of objects or environments does not depend on the camera pose, the estimated two sets of 3D point clouds in the same coordinates from different camera pose should be consistent.

Several previous studies have proposed various objectives that attempt to penalize inconsistencies in 3D geometric constraints~\cite{mahjourian2018unsupervised,gordon2019depth,fei2019geo,luo2020consistent}.
However, their performance is limited owing to a geometrically inconsistent evaluation with an approximation and the dependence on other undifferentiable algorithms to obtain the coupling between point clouds~\cite{mahjourian2018unsupervised,gordon2019depth,fei2019geo,luo2020consistent}.
In this work, we address these issues to improve the depth and pose estimation accuracy. Inspired by recent successful works on the Wasserstein distance for different tasks, such as generative modeling~\cite{arjovsky2017wasserstein}, embedding~\cite{huang2016supervised,courty2018learning}, and domain adaptation~\cite{courty2017optimal}, we extend these applications to depth and pose estimation.
The proposed method employs the Wasserstein distance as a pluggable objective to measure the consistency between two point clouds and penalize it for more accurate estimation.

In contrast to the baselines, our method attempts to measure a geometric consistency from 3D point clouds, without any indirect process and bold approximation.
In addition, the mathematical formulation of our objective is smooth and symmetric, which is advantageous for an efficient and effective training process.
\begin{figure}[t]
  \begin{center}
  \hspace*{-5mm}
      \includegraphics[width=0.55\hsize]{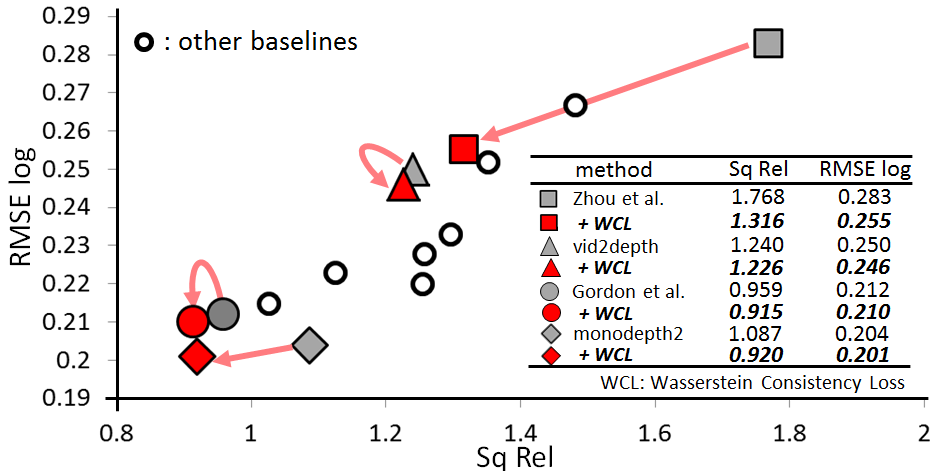}
  \end{center}
      \vspace*{-3mm}
	\caption{\small {\bf Evaluation of monocular depth estimation, with and without our proposed Wasserstein consistency loss (WCL) on the KITTI raw dataset, with an image size of 416$\times$128.} Red markers are the results with the WCL. Gray markers are the baselines without the WCL. Our WCL can enhance the performance of monocular depth estimation. It should be noted that a smaller Sq Rel and RMSE log are better. Mathematical formulation of Sq Rel and RMSE log are explained in supplemental material.}
  \label{f:pull_fig}
  \vspace*{-3mm}
\end{figure}
The major contributions of this study are 1) the proposal of a novel objective, the Wasserstein consistency loss~(WCL), to evaluate consistency in 3D geometric constraints, and 2) comparative evaluation to demonstrate that the accuracy of several state-of-the-art approaches is improved by ``plugging in'' the WCL. 
To the best of our knowledge, our work is the first application using the Wasserstein distance for depth and pose estimation.
We quantitatively and qualitatively demonstrate the benefits of our WCL using the KITTI dataset.

\section{Related Work}
\paragraph{Self-supervised monocular depth estimation}
Self-supervised depth estimation has recently become popularized ~\cite{zhou2017unsupervised,vijayanarasimhan2017sfm,yang2018every,ummenhofer2017demon,yang2018lego,mahjourian2018unsupervised,wang2018learning,casser2019depth,gordon2019depth,godard2019digging,fei2019geo,pillai2019superdepth,guizilini2020robust,guizilini2019packnet,patil2020don,johnston2020self,zhang2020online,garg2016unsupervised,godard2017unsupervised}.
Zhou et al.~\cite{zhou2017unsupervised} and Vijayanarasimhan et al.~\cite{vijayanarasimhan2017sfm} demonstrated one of the first approaches for self-supervised monocular depth and pose estimation.
Their approach simultaneously trained the model to estimate depth and pose based on knowledge of structure from motion (SfM).
Garg et al.~\cite{garg2016unsupervised} and Godard et al.\cite{godard2017unsupervised} proposed image reconstruction using SfM techniques between stereo images to estimate the depth, without pose estimation.

Several works were subsequently presented to improve accuracy.
Casser et al.~\cite{casser2019depth} introduced a motion mask based on semantic segmentation results to remove the dynamic object effect.
Godard et al.~\cite{godard2019digging} proposed monodepth2, which had a modified image reconstruction loss, considering occlusion.
CS Kumar et al.~\cite{cs2018depthnet} applied a recurrent neural network to understand the environment geometry from a video. 
Pillai et al.~\cite{pillai2019superdepth} demonstrated that a higher resolution input image can enable a more accurate estimation.
Guizilini et al.~\cite{guizilini2020robust} proposed image reconstruction in different camera poses to update the neural network in semi-supervised learning efficiently. 
In addition to these emergent techniques, the following section presents most works related to penalizing geometric inconsistencies between two point clouds estimated from different poses.

\paragraph{Penalization of Geometric Inconsistency}
Mahjourian et al.~\cite{mahjourian2018unsupervised} proposed a 3D point cloud alignment loss (iterative closest point (ICP) loss) to penalize the inconsistencies between two point clouds.
Their approach employs an ICP to form the coupling between two point clouds estimated from the images captured at different poses.
However, an ICP is not differentiable. Hence, the ICP loss separately computes the pose inconsistencies and the residual error between two point clouds, with approximation.

Gordon et al.~\cite{gordon2019depth} proposed the depth and ego-motion estimation system in the wild by learning the intrinsic camera parameters.
In addition, they introduced a depth consistency loss to minimize the difference between two estimated depth images from different frames.
They shared a bi-linear sampler for the image reconstruction loss~\cite{zhou2017unsupervised,jaderberg2015spatial} to determine the coupling and penalize any geometric inconsistency. As coupling is determined from the estimated depth itself, the positive effect of their depth consistency loss can be limited because the estimated depth will be indefinite.

Luo et al.~\cite{luo2020consistent} leveraged the optical flow to establish the coupling between point clouds and used these couplings for extracting 3D geometric constraints.
As the optical flow is estimated by a pre-trained neural network, the coupling performance largely depends on the domain of the dataset. It is reported that their approach would not work well on the KITTI dataset.

Fei et al.~\cite{fei2019geo} introduced a semantically informed geometric loss to penalize deviations from a horizontal or vertical plane.
They leveraged the results of the semantic segmentation from a pre-trained network and an inertial measurement sensor to determine if the segmented areas belonged to a horizontal or vertical plane. Although the surface of some objects can be an exact horizontal or vertical plane, e.g., a wall, and road, their method cannot support most objects with complex shapes.

\section{Proposed Method}
\begin{figure*}[t]
  \begin{center}
  \hspace*{-5mm}
      \includegraphics[width=0.95\hsize]{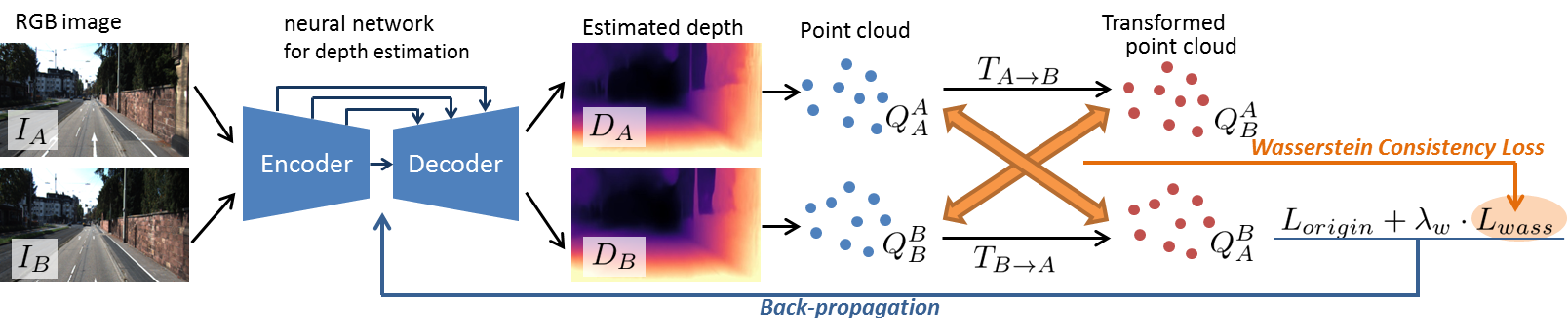}
  \end{center}
      \vspace*{-3mm}
	\caption{\small {\bf Overview of our proposed approach.} We feed RGB images $I_A$ and $I_B$ into a neural network to estimate depth images $D_A$ and $D_B$, respectively. From $D_A$ and $D_B$, we obtain the point clouds $Q_A^A$ and $Q_B^B$, and $Q_A^B$, and $Q_B^A$, respectively, by using the intrinsic camera parameters and the estimated transformation matrices, $T_{A \rightarrow B}$ and $T_{B \rightarrow A}$, respectively. To penalize a geometric inconsistency, we propose the addition of the WCL $L_{wass}$ to the original cost functions, $L_{origin}$, of recent state-of-the-art approaches; this is done to train the neural networks.}
  \label{f:overview}
\end{figure*}

\begin{figure}[t]
  \begin{center}
  \hspace*{-3mm}
    \includegraphics[width=0.8\hsize]{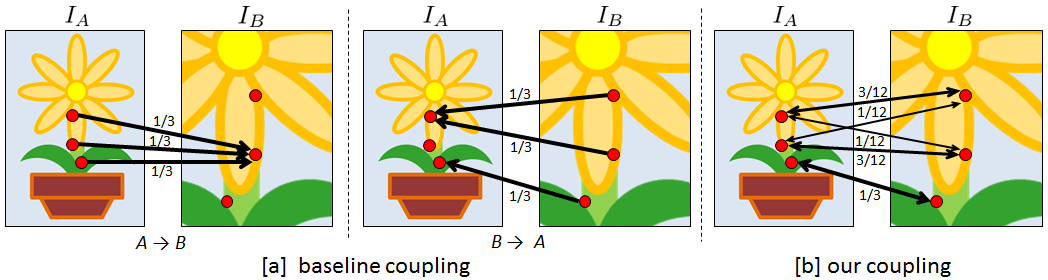}
  \end{center}
    \vspace{-3mm}
    \caption{\small {\bf Overview of the baseline and our coupling.} The three red points in each image indicate the position of the estimated depth in the image space. The arrows between images at pose A and B indicate the coupling example. The number on the arrow is the assigned mass, which is delivered to the coupled points. The thickness of the arrow visually displays its assigned mass. Our coupling with the Wasserstein distance is a soft coupling with the preservation law, which can completely receive and deliver the same constant value (=$1/m$=$1/n$=1/3). Our coupling is symmetric, which is different from the baselines.}
  \label{f:benefit}
\end{figure}

\subsection{Overview}
From previous studies, it is known that the penalization of a geometric inconsistency can enhance depth and pose estimation accuracy ~\cite{mahjourian2018unsupervised,gordon2019depth,fei2019geo,luo2020consistent}. 
Inspired by these approaches, we propose a novel pluggable WCL, shown as $L_{wass}$ in Fig.~\ref{f:overview}, for depth and pose estimation.
$L_{wass}$ works by adding $L_{origin}$ from the target method to the original objective, as shown in Fig.~\ref{f:overview}. In this paper, we focus on explaining $L_{wass}$, instead of the mathematical formulation of $L_{origin}$, which is the same as in prior works.
%
\paragraph{Preliminary} All prior works have commonly estimated the depth image as $D_A = f_{depth}(I_A)$, where $f_{depth}(\cdot)$ is a neural network for depth estimation, and $I_A$ is the image at camera pose $A$.
The estimated depth image, $D_A$, can be projected on to a 3D point cloud $Q_A^{A}$.
\begin{eqnarray}
    Q_A^A[i,j] = D_A[i,j] \cdot K^{-1} [i,j,1]^T
    \label{eq:pc}
\end{eqnarray}
Here, $i,j$ denotes the pixel position in the image coordinates, $K$ denotes an intrinsic camera parameter, and $Q_X^Y$ denotes a 3D point cloud at coordinate $X$ from $I_Y$.
It should be noted that $K$ is given as a constant matrix, except for the baseline~\cite{gordon2019depth}.
By calculating \eqref{eq:pc} for the entire image space, we obtain the point cloud $Q_A^A$ at coordinate $A$. 
By applying the same process to $I_B$, we can further obtain $Q_B^B$.
In addition, $Q_A^A$ and $Q_B^B$ can be transformed into mutual coordinates as $Q_B^A$ and $Q_A^B$, respectively, by multiplying with the transformation matrix $T_{A \rightarrow B}$ and $T_{B \rightarrow A}$, respectively.
$T_{A \rightarrow B}$ and $T_{B \rightarrow A}$ are estimated from $I_A$ and $I_B$, respectively, through another neural network.
$(Q_A^A, Q_A^B)$ and $(Q_B^B, Q_B^A)$ are sets of estimated point clouds at coordinates A and B, respectively.

\paragraph{Overview of WCL} Similar to~\cite{mahjourian2018unsupervised,gordon2019depth,luo2020consistent}, our WCL penalizes the inconsistency between two point clouds at the same coordinate. $L_{wass}$ is defined as follows:
\begin{eqnarray}
    L_{wass} = \wassapprox(Q_A^A, Q_A^B) + \wassapprox(Q_B^B, Q_B^A).
    \label{eq:Lwass}
\end{eqnarray}
Here, $\wassapprox(\cdot)$ is the (squared) Wasserstein distance between two point clouds, as is subsequently described.
We add $\lambda_w \cdot L_{wass}$ to the original objective, $L_{origin}$, and update the neural networks for depth and pose estimation by minimizing the combined objective.
$\lambda_w$ is a weighting factor of $L_{wass}$ to balance $L_{origin}$.
The benefits of our WCL are,
\begin{enumerate}
  \item geometrically consistent penalization between 3D point clouds,
  \item smooth and symmetric objective,
  \item simple implementation.
\end{enumerate}
In the baselines~\cite{mahjourian2018unsupervised,gordon2019depth,luo2020consistent}, the geometric consistency between two point clouds is calculated in two steps: 1) taking a coupling between two point clouds and 2) calculating the distance(e.g., Euclidean distance) between the aligned point clouds using the coupling. 

Fig.~\ref{f:benefit} shows a simple example of a coupling using three points ($m=n=3$)  on each image. 
Here, $m$ and $n$ are the size of two point clouds estimated from $I_A$ and $I_B$, respectively. 
In the baseline coupling~\cite{mahjourian2018unsupervised,gordon2019depth,luo2020consistent}, multiple geometrically different points on one image can correspond to one point on the other image and the coupling of $A \rightarrow B$ and $B \rightarrow A$ can be asymmetric, as shown in Fig.~\ref{f:benefit}[a]. 
These behaviors are geometrically inconsistent.
Furthermore, the baselines cause discontinuous coupling between each training iteration as updating a neural network for depth estimation changes the coupling, and the coupling itself is a hard coupling, that one point on $I_A$ can make a coupling only with one point on $I_B$ for $A \rightarrow B$ coupling. The same is true for $B \rightarrow A$ coupling.
It means that the whole assigned mass, $\frac{1}{3} ( = \frac{1}{m} = \frac{1}{n})$, is delivered to exactly one point. Here, ``mass'' refers to the weight equally assigned to all points and is a term mainly used in optimal transport~\cite{peyre2019computational}.

Our proposed WCL can address these issues to improve the depth and pose estimation accuracy. 
In contrast to the baselines, our coupling with the Wasserstein distance is a soft coupling in which the mass assigned is preserved, as shown in Fig.~\ref{f:benefit}[b]. The sum of the delivered and received values are the same as the assigned mass $\frac{1}{3}$.
Moreover, our soft coupling by the Wasserstein distance is symmetric, as shown in Fig.~\ref{f:benefit}[b].
Hence, our WCL can enforce avoidance of geometrically inconsistent penalization. 
In addition, our WCL can simultaneously 1) take a soft coupling and 2) calculate the distance between two point clouds.
These processes of our WCL are totally differentiable without involving other external libraries (e.g., ICP) and results in a more stable learning compared to non-differentiable baselines~\cite{mahjourian2018unsupervised,fei2019geo,luo2020consistent}.
As a result, this entire algorithm can be implemented only on a GPU. Furthermore, the code size of our WCL is relatively short, as shown later.
%
\subsection{WCL}
We introduce the \emph{Wasserstein distance}~\cite{peyre2019computational} and its approximation $\wassapprox(\cdot)$ used in $L_{wass}$, which provides a differentiable loss function that penalizes the geometric inconsistency between two point clouds.

\paragraph{Preliminary: Definition and Notation}
In this subsection, we express the point clouds as ${\cal X}=\{x_1,\cdots,x_m\}$ and ${\cal Y}=\{y_1,\cdots,y_n\}$, to simplify and generalize the mathematical formulation. 
Here, ${\cal X}$ and ${\cal Y}$ are a set of points in $\mathbb{R}^3$.
%
$\langle \vec{U}, \vec{V}\rangle$ denotes the inner product between two vectors (or matrices) of the same size. $\onevec{n}$ is an $n$-dimensional vector, whose elements are all one.

\begin{figure}
    \centering
    \includegraphics[width=0.6\linewidth]{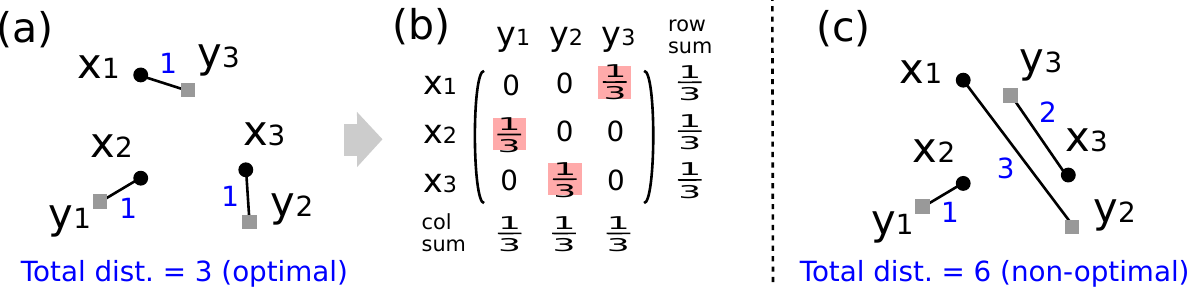}
    \caption{{\bf Wasserstein distance.} (a) Optimal coupling; (b) Matrix $\vec{P}$ corresponding to the coupling displayed in (a); (c) Non-optimal coupling; The blue numbers are the squared distances between two points. It should be noted that this figure is simplified; we can, in fact, treat soft coupling, which distributes the weight from one to many (see $\mathcal{U}_{m,n}$). }
    \label{fig:wass}
\end{figure}

\paragraph{Wasserstein Distance}
%
The \emph{Wasserstein distance}\footnote{Originally, the Wasserstein distance was defined for probabilistic distributions. Currently, point clouds can be regarded as mixtures of delta functions; hence, we can deal with point clouds using the Wasserstein distance.} is a metric between two point clouds. 
This is defined as the sum of the squared Euclidean distances between two ``coupled'' points in ${\cal X}$ and ${\cal Y}$.
An example is shown in Fig.~\ref{fig:wass}(a).
Suppose two point clouds ${\cal X}=\{x_1, x_2, x_3\}$ and ${\cal Y}=\{y_1, y_2, y_3\}$, and consider a coupling $(x_1, y_3)$, $(x_2, y_1)$, $(x_3, y_2)$.
This coupling clearly minimizes the total distance between coupled points; no other coupling can reduce the total distance (e.g., the coupling in Fig.~\ref{fig:wass}(c)).
However, such an optimal coupling is unknown in advance; therefore, we need to determine the optimal coupling that minimizes the total distance (e.g., a non-optimal coupling in Fig.~\ref{fig:wass}(c) does not provide the valid Wasserstein distance).
Essentially, identifying such an optimal coupling corresponds to the following optimization problem~\cite{peyre2019computational}:
\begin{align}
    &\min_{\vec{P}}\; \langle \vec{P}, \vec{C}\rangle
    \quad
    \text{subject to}\; \vec{P}\in \mathcal{U}_{m,n}.\label{eq:wassdef}
\end{align}

Here, $\vec{C}$ is an $m$-by-$n$ Euclidean distance matrix whose $(i,j)$ element is the squared distance between $x_i$ and $y_j$ (i.e., $\vec{C}_{ij}=\|x_i-y_j\|^2$), and $\vec{P}$ is an $m$-by-$n$ coupling matrix whose $(i,j)$ element has amount of mass delivered from $x_i$ to $y_j$.
Moreover, $\mathcal{U}_{m,n}$ is a set of $m$-by-$n$ matrices that represent a valid coupling, formally defined by
\begin{align}
    \mathcal{U}_{m,n}=\{\vec{P}\in\mathbb{R}_{\ge0}^{m\times n}\mid \vec{P}\onevec{n}=\frac{\onevec{m}}m, \vec{P}^\top\onevec{m}=\frac{\onevec{n}}n\}.
    \label{eq:wassconst}
\end{align}
This suggests that the mass assigned to each point in ${\cal X}$ (i.e., $\frac1m$) must be delivered to points in ${\cal Y}$ \emph{without overs and shorts}, and \textit{vice versa}; this conservation law is entirely different to the existing geometric inconsistency losses.
The example in Fig.~\ref{fig:wass}(b) represents the coupling matrix corresponding to Fig.~\ref{fig:wass}(a).
Elements of $\vec{P}$ corresponding to the coupled points (i.e., $(x_1, y_3)$, $(x_2, y_1)$, $(x_3, y_2)$) are filled with $\frac13$, whereas the others are zero.
This $\vec{P}$ satisfies the constraint \eqref{eq:wassconst}; the sums of each row and column are $\frac1m$ and $\frac1n$, respectively (it should be noted that we have $m=n=3$ here).

With the optimal solution $\vec{P}^*$, the Wasserstein distance is defined by $W^2({\cal X},{\cal Y})= \langle\vec{P}^*,\vec{C}\rangle$.
It is known that $W^2(\cdot, \cdot)$ defines a proper metric, i.e., the Wasserstein distance satisfies three axioms\footnote{That is, $W({\cal X},{\cal Y})$ satisfies: (i) $W({\cal X},{\cal Y})=0\Leftrightarrow {\cal X}={\cal Y}$ (identity of indiscernibles); (ii) $W({\cal X},{\cal Y})=W({\cal Y},{\cal X})$ (symmetry); (iii) $W({\cal X},{\cal Y})\le W({\cal X},{\cal Z})+W({\cal Z},{\cal Y})$ (triangle inequality).} of metric~\cite{peyre2019computational}; hence, we expect it to behave properly as a loss function.
Note that $\vec{P}$ provides soft coupling to allow weighted coupling with multiple points, although Fig.\ref{fig:wass} shows the simplified case with only one-to-one coupling; we can define \eqref{eq:wassdef}, even if $m\ne n$, i.e., the size of ${\cal X}$ and ${\cal Y}$ are different.

%
%
\paragraph{Computing WCL and its Gradient}
We train the neural networks by reducing the geometric inconsistency of two point clouds measured by the Wasserstein distance.
Thus, we need to compute $W^2({\cal X},{\cal Y})$ and its gradients with respect to ~${\cal X}$ and ${\cal Y}$.
\emph{Sinkhorn iteration} (Algorithm~\ref{algo:sinkhorn}) allows us to compute $W^2({\cal X},{\cal Y})=\langle \vec{P}^*,\vec{C}\rangle$ very accurately, as well as its gradients.
Benefits of the Sinkhorn iteration are two-fold:
(1) \emph{we can use GPUs} as it combines simple arithmetic operations;
(2) \emph{we can back-prop the iteration directly} through auto-gradient techniques equipped in most modern deep learning libraries (i.e., we do not need external libraries, unlike the ICP loss).

\begin{algorithm}[htbp]
  \caption{Computing WCL $\wassapprox({\cal X},{\cal Y})$ by the Sinkhorn iteration. $\varepsilon>0$ is a small constant.}\label{algo:sinkhorn}
  \DontPrintSemicolon
  \KwData{Point clouds ${\cal X}=\{x_i\}_{i=1}^m$, ${\cal Y}=\{y_j\}_{j=1}^n$}
  \nl $\vec{C}_{ij}=\|x_i-y_j\|^2 \qquad 1\le \forall i\le m, \; 1\le\forall j\le n$\;
  \nl $\vec{G}\leftarrow\exp(-\vec{C}/\varepsilon)$\tcp*{\footnotesize{Element-wise exp}}
  \nl $\vec{v}\leftarrow \onevec{n}$\tcp*{\footnotesize{Initialize dual variable}}
  \nl \While{Not converged}{
      \nl $\vec{u}\leftarrow \frac1m\onevec{m}/\vec{G}\vec{v}$\tcp*{\footnotesize{Element-wise division}}
      \nl $\vec{v}\leftarrow \frac1n\onevec{n}/\vec{G}^\top\vec{u}$
  }
  \nl \textbf{{return}} \, {$\langle\diag{\vec{u}}\vec{G}\,\diag{\vec{v}},\vec{C}\rangle$ \textbf{as} $\wassapprox({\cal X},{\cal Y})$}\;
\end{algorithm}

\paragraph{Remark}
To be exact, Algorithm~\ref{algo:sinkhorn} is an approximation algorithm for \eqref{eq:wassdef};
it solves the optimization problem with an additional \emph{entropy regularization} term $H(\vec{P})$:
\begin{align}
    &
    \min_{\vec{P}} \langle \vec{P},\vec{C}\rangle-\varepsilon H(\vec{P}), \quad \text{subject to} \; \vec{P}\in\mathcal{U}_{m,n}
    \label{eq:regwass}\\
    &
    \text{where}\quad H(\vec{P}):=-\sum_{ij}\vec{P}_{ij}(\log\vec{P}_{ij}-1).\nonumber
\end{align}
With the optimal solution $\vec{P}^\dagger$ of this problem, we define the regularized Wasserstein distance by $\wassapprox({\cal X},{\cal Y})=\langle\vec{P}^\dagger,\vec{C}\rangle.$
This regularization term makes WCL smooth with respect to its inputs, resulting in stable training.
As can be seen, if $\varepsilon\to0$, \eqref{eq:regwass} converges to the original optimization problem \eqref{eq:wassdef};
hence, a small $\varepsilon>0$ gives a good approximation. However, such a small $\varepsilon$ might cause numerical instability as a small $\varepsilon$ reduces $\vec{G}$ to an almost zero matrix at Line~2. 
To improve numerical stability, we may use log-sum-exp at Lines 5 $-$ 6.
Furthermore, we can compute Algorithm~\ref{algo:sinkhorn} in parallel over batch dimension.
At Line~4, we can use any kind of stopping condition; here, we stop at 100 iterations.
See \cite{cuturi2013sinkhorn,peyre2019computational} for details of the regularized Wasserstein distance.

\section{Experiment}
We apply our WCL to prior works and quantitatively and qualitatively evaluate them using the public dataset with comparison to the original method.

\subsection{Dataset}
In this study, we use the KITTI dataset~\cite{Geiger2013IJRR}.
Although it is known that a large input image size can lead to better performance~\cite{pillai2019superdepth}, the main purpose of our experiment is to confirm any advantages against the target original method to be applied.
Hence, we use a 416$\times$128 image size, similar to prior works~\cite{zhou2017unsupervised,mahjourian2018unsupervised,gordon2019depth,godard2019digging}.
However, we conduct an additional evaluation using an image size of 640$\times$192, for monodepth2 only ~\cite{godard2019digging} as their default image size is 640$\times$192 in the public code. As with several prior works, we have two different datasets, split for the evaluation of depth and pose estimation. It should be noted that the models for depth and pose estimation are trained simultaneously on each dataset, but depth and pose are trained and evaluated on separate data.
\paragraph{Depth Estimation} We separate the KITTI raw dataset via the Eigen split~\cite{eigen2014depth}. As a result, we have approximately 40,000 frames for training, 4,000 frames for validation, and 697 frames for the test.
The test data are chosen from 29 scenes of the KITTI raw dataset. 
Although the KITTI raw data include stereo images and LIDAR data, we use monocular images for both training and testing as the input data and use the LIDAR data as the ground truth only for testing.
\paragraph{Pose Estimation} Unlike the KITTI raw dataset, the KITTI odometry dataset has the ground truth of relative pose between consecutive images for testing.
However, test data in the KITTI odometry dataset are partially included in the training data of the KITTI raw dataset.
Hence, we can not use the trained model on the KITTI raw dataset for the pose estimation evaluation on the KITTI odometry dataset.
Therefore, we have both training and testing on the KITTI odometry dataset, as with the baseline methods~\cite{zhou2017unsupervised,godard2019digging}. 
Although there are 22 sequences in the KITTI odometry dataset, we use 11 sequences, from 00 to 10. 
After training models on sequence 00 to 08, we test the model for pose estimation on sequence 09 and 10.

\subsection{Evaluation of Depth Estimation}
\paragraph{Training}
In the evaluation of depth estimation, we apply our WCL to four selected baselines~\cite{zhou2017unsupervised,mahjourian2018unsupervised,gordon2019depth,godard2019digging}.
For \cite{zhou2017unsupervised,godard2019digging}, we add $\lambda_{w} \cdot L_{wass}$ to their original cost function, $L_{origin}$, to train the model.
By contrast, as the original cost function of \cite{mahjourian2018unsupervised,gordon2019depth} includes the ICP loss~\cite{mahjourian2018unsupervised} and the depth consistency loss\cite{gordon2019depth}, which also penalizes geometric inconsistencies, we remove them for the comparative evaluation with each objective.
We train all models by applying the same hyperparameters (e.g., mini-batch size, learning rate, data augmentation, and network structure) and training process (e.g., masking, training length, optimization, selection of input images $I_A$ and $I_B$ in Fig.~\ref{f:overview}) as in the original code, except for the following hyperparameters about our WCL.

We determine the weighting values, $\lambda_{w}$, for $L_{wass}$ as 7.0, 2.0, 3.0, and 0.5 in \cite{zhou2017unsupervised}, \cite{mahjourian2018unsupervised}, \cite{gordon2019depth}, and \cite{godard2019digging}, respectively.
$\lambda_w$ is a weighting factor that balances $L_{origin}$; therefore, there is no significance to the relative size of the values. 
$\varepsilon$ in \eqref{eq:regwass} is set as 0.001 to suppress the approximation and to stably calculate the WCL. 
In addition, we uniformly sample the point clouds at a grid point on an image coordinate before feeding them into our WCL in \eqref{eq:Lwass} owing to the limitation of GPU memory usage. 
An explanation of the sampling method and an ablation study of the hyperparameters are shown in the supplementary materials.


\paragraph{Quantitative Analysis}
Table~\ref{tab:ev} displays the widely used seven metrics for the evaluation of depth estimation from monocular camera images. 
As summarized in Table~\ref{tab:ev}, the performance of all the baselines can be successfully improved on most metrics by adding our WCL.
The improvement for \cite{zhou2017unsupervised} and \cite{godard2019digging} is about 25$\%$ and 15$\%$ on the most advantageous metric, respectively. On the other hand, it is not quite as large for \cite{mahjourian2018unsupervised} and \cite{gordon2019depth} as they also penalize geometric inconsistencies using their own approach.
However, our method still has explicit margins about 5$\%$ on Sq Rel or RMSE against these baselines.
Furthermore, monodepth2~\cite{godard2019digging} + WCL can successfully obtain the best results, which are quite better than the other related works~\cite{luo2020consistent,fei2019geo}.

\begin{table*}[t]
  \caption{{\small {\bf Evaluation of depth estimation by self-supervised mono supervision on an Eigen split of the KITTI dataset.} We display seven metrics from the {\bf estimated depth images less than 80 m}. ``+ WCL'' is the result obtained with our proposed method, in addition to the baseline method presented above. For the leftmost four metrics, smaller is better; for the rightmost three metrics, higher is better. $\dagger$ and $\ddagger$ indicate removing the ICP and depth consistency loss from the original cost function, respectively. The method of $\ddagger$ is evaluated by the author. The $\ast$ method uses the ground truth pose.}}
  \begin{center}
  \resizebox{0.95\columnwidth}{!}{
  \label{tab:ev}
  \begin{tabular}{|lc|c|c|c|c||c|c|c|} \hline
    method & image size & Abs Rel & Sq Rel & RMSE & RMSE log & $\delta < 1.25$ & $\delta < 1.25^2$ & $\delta < 1.25^3$ \\ \hline
    Yang et al.~\cite{yang2017unsupervised} & 416x128 & 0.182 & 1.481 & 6.501 & 0.267 & 0.725 & 0.906 & 0.963 \\
    LEGO~\cite{yang2018lego} & 416x128 & 0.162 & 1.352 & 6.276 & 0.252 & 0.783 & 0.921 & 0.969 \\
    GeoNet~\cite{yin2018geonet} & 416x128 & 0.155 & 1.296 & 5.857 & 0.233 & 0.793 & 0.931 & 0.973 \\
    Fei et al.~\cite{fei2019geo} & 416x128 & 0.142 & 1.124 & 5.611 & 0.223 & 0.813 & 0.938 & 0.975 \\
    DDVO~\cite{wang2018learning} & 416x128 & 0.151 & 1.257 & 5.583 & 0.228 & 0.810 & 0.936 & 0.974 \\
    Yang et al.~\cite{yang2018every} & 416x128 & 0.131 & 1.254 & 6.117 & 0.220 & 0.826 & 0.931 & 0.973 \\
    Casser et al.~\cite{casser2019depth} & 416x128 & 0.141 & 1.026 & 5.291 & 0.2153 & 0.8160 & 0.9452 & 0.9791 \\
    Luo et al.~\cite{luo2020consistent} $\ast$ & 384x112 & 0.130 & 2.086 & 4.876 & 0.205 & 0.878 & 0.946 & 0.970 \\ \hline
    Zhou et al.~\cite{zhou2017unsupervised} & 416x128 & 0.208 & 1.768 & 6.856 & 0.283 & 0.678 & 0.885 & 0.957 \\
    \bf{+ WCL} & 416x128 & \bf{0.171} & \bf{1.316} & \bf{6.080} & \bf{0.255} & \bf{0.755} & \bf{0.915} & \bf{0.966} \\ \hdashline
    vid2depth~\cite{mahjourian2018unsupervised} & 416x128 & \bf{0.163} & 1.240 & 6.220 & 0.250 & 0.762 & 0.916 & \bf{0.968} \\
    vid2depth~\cite{mahjourian2018unsupervised} {\scriptsize wo ICP loss}$\dagger$ & 416x128 & 0.175 & 1.617 & 6.267 & 0.252 & 0.759 & 0.917 & 0.967 \\
    \bf{+ WCL} & 416x128 & 0.165 & \bf{1.226} & \bf{5.892} & \bf{0.246} & \bf{0.767} & \bf{0.918} & \bf{0.968} \\ \hdashline
    Gordon et al.~\cite{gordon2019depth} & 416x128 & 0.128 & 0.959 & 5.23 & 0.212 & \bf{0.845} & \bf{0.947} & 0.976 \\
    Gordon et al.~\cite{gordon2019depth} {\scriptsize wo depth consis. loss}$\ddagger$ & 416x128 & 0.129 & 0.945 & \bf{5.211} & 0.214 & 0.839 & 0.944 & 0.976 \\    
    \bf{+ WCL} & 416x128 & \bf{0.125} & \bf{0.915} & 5.231 & \bf{0.210} & 0.844 & \bf{0.947} & \bf{0.977} \\ \hdashline
    monodepth2~\cite{godard2019digging} & 416x128 & 0.128 & 1.087 & 5.171 & 0.204 & 0.855 & \bf{0.953} & 0.978 \\
    \bf{+ WCL} & 416x128 & \bf{0.123} & \bf{0.920} & \bf{4.990} & \bf{0.201} & \bf{0.858} & \bf{0.953} & \bf{0.980} \\  \hdashline
    monodepth2~\cite{godard2019digging} & 640x192 & 0.115 & 0.903 & 4.863 & 0.193 & \bf{0.877} & \bf{0.959} & \bf{0.981} \\
    \bf{+ WCL} & 640x192 & \bf{0.114} & \bf{0.813} & \bf{4.705} & \bf{0.191} & 0.874 & \bf{0.959} & \bf{0.981} \\ \hline     
  \end{tabular}
  }
\end{center}
  \vspace{-5mm}
\end{table*}

\begin{figure*}[t]
  \begin{center}
      \includegraphics[width=0.99\hsize]{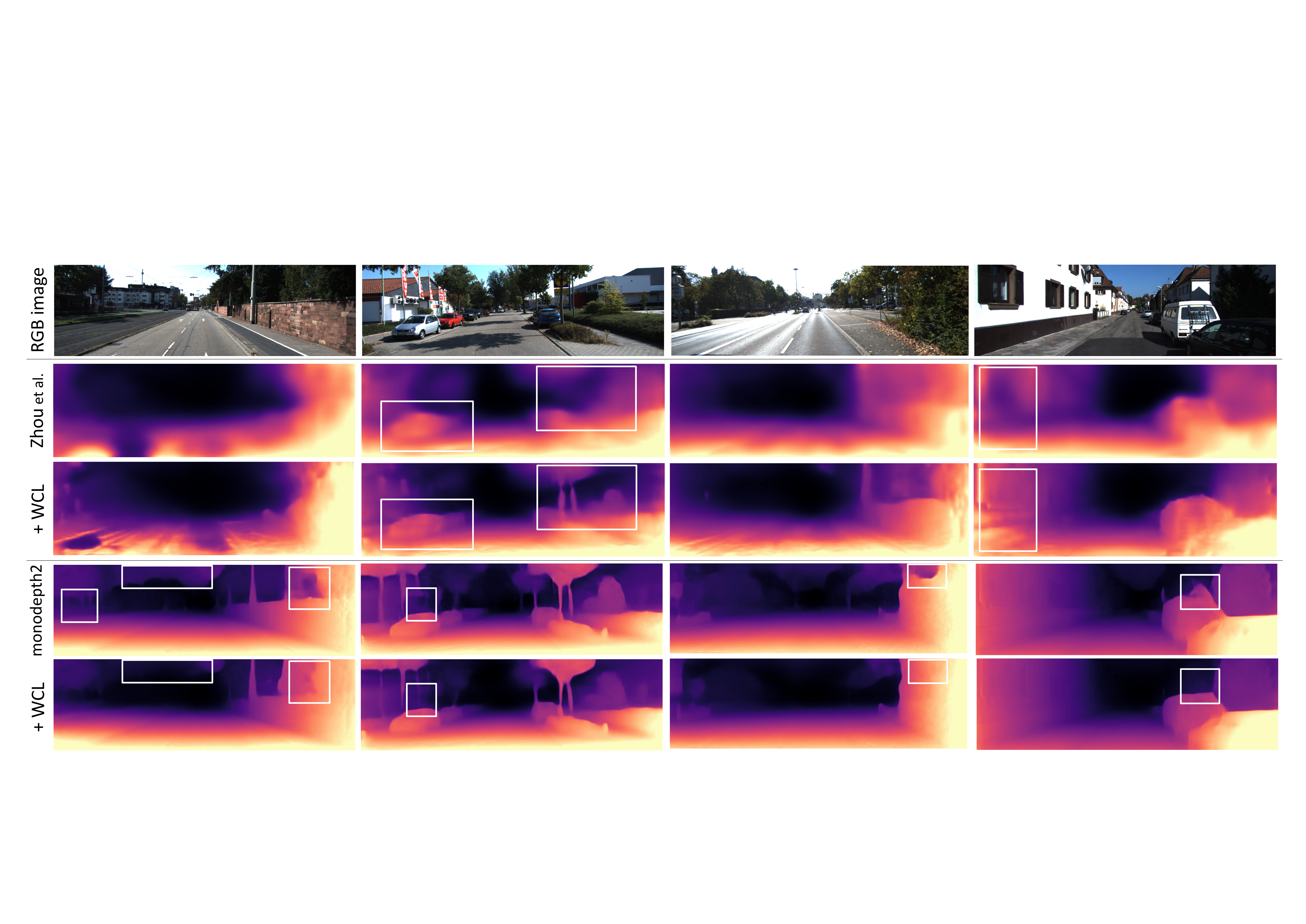}
  \end{center}
	\caption{\small {\bf Qualitative results of our proposed method.} The top row displays the input images for the trained neural network to estimate depth images and the other rows display the estimated depth images, with and without the WCL. The white rectangle box in the depth image highlights the advantages of the WCL.}
  \label{f:pred_depth}
\end{figure*}

\paragraph{Qualitative Analysis}
We show the depth images of \cite{zhou2017unsupervised,godard2019digging}, with and without our WCL, in Fig.~\ref{f:pred_depth}.
The first row in Fig.~\ref{f:pred_depth} shows the RGB image as the neural network input, the second and fourth rows are the depth images estimated by \cite{zhou2017unsupervised} and \cite{godard2019digging}, and the third and fifth rows are the depth images estimated by \cite{zhou2017unsupervised} + WCL and \cite{godard2019digging} + WCL, respectively.
In the estimated depth images, we display a white lined rectangle to highlight the advantage of our method.
Our method can reduce the number of artifacts and sharpen estimation.
The depth images in additional cases estimated by various methods in Table~\ref{tab:ev} are shown in the supplementary material.

\subsection{Evaluation of Pose Estimation}
For evaluation of pose estimation, we select two prior works~\cite{zhou2017unsupervised,godard2019digging} to apply our WCL.
We could not evaluate \cite{mahjourian2018unsupervised,gordon2019depth} for pose estimation, because the hyperparameters and codes for the pose estimation and evaluation are partially unclear from their released code. 
For \cite{zhou2017unsupervised,godard2019digging}, we train the models using the KITTI odometry dataset, with the same training conditions as the depth estimation.

Table~\ref{tab:ego} displays the absolute trajectory error (ATE) in meters for sequence 09 and 10 of the KITTI odometry dataset.
We show the mean and standard deviation of ATE over all overlapping 5 frame snippets. 
By applying our WCL, the pose estimation accuracies of two prior studies~\cite{zhou2017unsupervised,godard2019digging} are improved.
Although the performance is slightly worse than ``ORB-SLAM (full)'', the original ORB-SLAM with loop closure using whole sequence images, the methods with our WCL outperforms ``ORB-SLAM (short)'', which takes 5 consecutive images like our method.

\begin{table}[t]
  \caption{{\small {\bf Evaluation of pose estimation on sequence 09 and 10 of the KITTI odometry dataset.} Results show the mean and standard deviation of the absolute trajectory error in meters. ``frame number'' is the number of input images used for pose estimation. The method with our WCL outperforms the original method.}}
  \begin{center}
  \resizebox{0.75\columnwidth}{!}{
  \label{tab:ego}
  \begin{tabular}{|lc|c|c|c|} \hline
    method & image size & \hspace{5mm} sequence 09 \hspace{5mm} & \hspace{5mm} sequence 10 \hspace{5mm} & frame number \\ \hline
    ORB-SLAM (full) & full size & 0.014 $\pm$ 0.008 & 0.012 $\pm$ 0.011 & - \\
    ORB-SLAM (short) & full size & 0.064 $\pm$ 0.141 & 0.064 $\pm$ 0.130 & - \\ \hline
    Zhou et al.~\cite{zhou2017unsupervised} & 416x128  & 0.021 $\pm$ 0.017 & 0.020 $\pm$ 0.015 & 5 \\
    \bf{+ WCL} & 416x128  & \bf{0.016} $\pm$ \bf{0.011} & \bf{0.013} $\pm$ \bf{0.009} & 5 \\ \hdashline
    monodepth2~\cite{godard2019digging} & 416x128  & 0.017 $\pm$ \bf{0.009} & 0.015 $\pm$ \bf{0.010} & 2\\
    \bf{+ WCL} & 416x128  & \bf{0.016} $\pm$ \bf{0.009} & \bf{0.014} $\pm$ \bf{0.010} & 2 \\  \hdashline
    monodepth2~\cite{godard2019digging} & 640x192  & 0.017 $\pm$ \bf{0.008} & 0.015 $\pm$ \bf{0.010} & 2 \\
    \bf{+ WCL} & 640x192  & \bf{0.016} $\pm$ \bf{0.008} & \bf{0.014} $\pm$ \bf{0.010} & 2\\ \hline     
  \end{tabular}
  }
\end{center}
  \vspace{-0.5em}
\end{table}

\section{Conclusion}
In this paper, we proposed a novel WCL to penalize geometric inconsistencies, for depth and pose estimation.
Our proposed approach employed the Wasserstein distance for measuring the consistency between two point clouds from different frames.
Our WCL is a smooth and symmetric objective, which can suitably measure geometric consistency without using any other external and/or non-differentiable libraries.
Therefore, the neural network can be effectively and efficiently trained to obtain highly accurate depth estimation.
In the experiment, we applied our proposed WCL to several state-of-the-art baselines and confirmed the benefits of our method with healthy margins.

There are two remaining issues to be addressed in the future.
First, the study of memory-saving methods for measuring geometric inconsistency.
As shown in the supplementary material, the performance of our WCL is restricted by the limitation on GPU memory.
Second, the occlusion when calculating Wasserstein distance.
Relaxation of the coupling constraints and/or masking for occluded point clouds will be required for more accurate estimation.
Our WCL has potentials to adapt for this issue, because our WCL can be calculated for pairs of different number of point clouds as $m \neq n$. 

\clearpage
\bibliography{egbib}  

\clearpage
\section{Supplementary Material}
\input{7-supplemental.tex}
\end{document}

%% file: packages.tex
\usepackage[ruled,vlined]{algorithm2e}
\usepackage{bbm}

%% file: macro.tex
\renewcommand{\vec}[1]{\boldsymbol{\textrm{#1}}}

\newcommand{\onevec}[1]{{\textbf 1}_{#1}}
\newcommand{\diag}[1]{\textrm{diag}(#1)}

\newcommand{\wassapprox}{\widetilde W^2}

%% file: 7-supplemental.tex
\subsection{Sparse Sampling}
\begin{figure}[h]
  \begin{center}
  \hspace*{-5mm}
      \includegraphics[width=0.7\hsize]{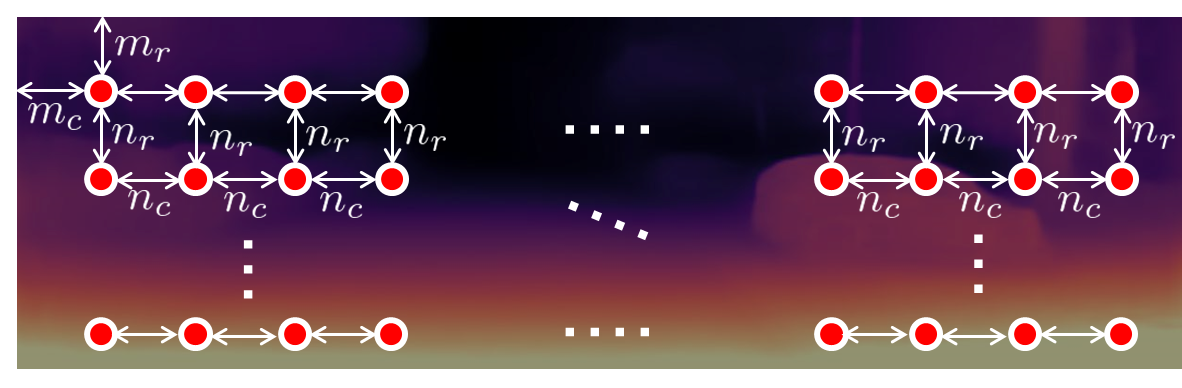}
  \end{center}
      \vspace*{-3mm}
	\caption{\small {\bf Sparse sampling of point clouds.} Red dots are the position of sampled point clouds in the image coordinates.}
  \label{f:sample}
  \vspace*{-3mm}
\end{figure}

In training, we need to reduce the number of the point cloud for calculation of the WCL owing to limited GPU memory usage.
We uniformly sample the point clouds on the depth image, as shown in Fig.~\ref{f:sample}.
Here, $n_c$ and $n_r$ indicate vertical and horizontal grid point intervals, respectively.
$m_c$ and $m_r$ indicate random offsets less than and equal to $n_c$ and $n_r$, respectively.
We cover the whole point clouds in training by randomly choosing $m_c$ and $m_r$ in each training iteration.

\subsection{Metrics of Depth Evaluation}
In quantitative analysis, we evaluate the estimated depth using seven metrics.
``Abs Rel'', ``Sq Rel'', ``RMSE'', and ``RMSE log'' in Table~\ref{tab:ev} are the means of the following values in the entire test data. 
\begin{itemize}
  \item Abs Rel\hspace{9mm}:\hspace{2mm} $|D_{gt} - \hat{D}|/D_{gt}$
  \item Sq Rel\hspace{11mm}:\hspace{2mm} $(D_{gt} - \hat{D})^2/D_{gt}$
  \item RMSE\hspace{11mm}:\hspace{2mm} $\sqrt{(D_{gt} - \hat{D})^2}$
  \item RMSE log\hspace{6mm}:\hspace{2mm} $\sqrt{(\mbox{log}(D_{gt}) - \mbox{log}(\hat{D}))^2}$
\end{itemize}
Here, $D_{gt}$ denotes the ground truth of the depth image and $\hat{D}$ denotes estimated depth. The remaining three metrics are the ratios to satisfy with $\delta < \alpha$. $\delta$ is calculated as follows:
\begin{eqnarray}
    \delta = \mbox{max}(D_{gt}/\hat{D}, \hat{D}/D_{gt}).
    \label{eq:delta}
\end{eqnarray}
As with prior works, we have three $\alpha$ as 1.25, 1.25$^2$ and 1.25$^3$.

\subsection{Ablation Study}
Table~\ref{tab:ab} shows the results of the ablation study in monodepth2~\cite{godard2019digging} with our WCL; the upper part is for weighting factor $\lambda_{w}$ and the lower part is for $n_c$ and $n_r$ of sparse sampling. At the top, $\lambda_{w}$ is changed under fixed $n_c$ and $n_r$; at the bottom, $n_c$ and $n_r$ are changed under fixed $\lambda_{w}$. From the top, it can be seen that increasing the weighting factor can improve the depth estimation performance but making it too large worsens the performance; in this case, the best result is obtained at $\lambda_{w}=0.5$.

In the bottom part, better performance can be obtained at smaller $n_c$ and $n_r$ as geometric inconsistencies can be accurately penalized. However, as it is difficult to make $n_c$ and $n_r$ even smaller due to the GPU memory limitations, we decide $n_c=16$ and $n_r=4$ for the evaluation in Table~\ref{tab:ev}.
\begin{table*}[h]
  \caption{{\small {\bf Ablation study on weighting value $\lambda_w$(top) and sparse sampling parameters $n_c$ and $n_r$(bottom)} for monodepth2~\cite{godard2019digging} + WCL. All metrics for evaluation are the  same as in Table~\ref{tab:ev}. Input and output resolutions are 416$\times$128.}}
  \begin{center}
  \resizebox{1.0\columnwidth}{!}{
  \label{tab:ab}
  \begin{tabular}{|lccc|c|c|c|c||c|c|c|} \hline
    method & $\lambda_w$ & $n_c$ & $n_r$ & Abs Rel & Sq Rel & RMSE & RMSE log & $\delta < 1.25$ & $\delta < 1.25^2$ & $\delta < 1.25^3$ \\ \hline
    monodepth2~\cite{godard2019digging} & $-$ & $-$ & $-$ & 0.128 & 1.087 & 5.171 & 0.204 & 0.855 & 0.953 & 0.978 \\
    \bf{+ WCL} & 0.1 & 16 & 4 & 0.127 & 0.995 & 5.039 & 0.204 & 0.853 & \bf{0.954} & 0.979 \\ 
    & 0.3 & 16 & 4 & 0.125 & 0.965 & 5.019 & 0.202 & 0.856 & 0.953 & 0.979 \\  
    & 0.5 & 16 & 4 & \bf{0.123} & \bf{0.920} & \bf{4.990} & \bf{0.201} & \bf{0.858} & 0.953 & \bf{0.980} \\
    & 0.7 & 16 & 4 & 0.124 & 0.922 & 5.023 & \bf{0.201} & 0.853 & \bf{0.954} & \bf{0.980} \\
    & 1.0 & 16 & 4 & 0.128 & 0.993 & 5.158 & 0.205 & 0.847 & 0.950 & 0.979 \\ 
    & 2.0 & 16 & 4 & 0.132 & 1.045 & 5.201 & 0.207 & 0.845 & 0.949 & 0.978 \\ \hdashline 
    \bf{+ WCL} & 0.5 & 64 & 16 & 0.127 & 0.961 & 5.143 & 0.204 & 0.849 & 0.952 & 0.979 \\
    & 0.5 & 32 & 16 & 0.127 & 0.976 & 5.052 & 0.203 & 0.852 & \bf{0.953} & 0.979 \\ 
    & 0.5 & 64 & 8 & 0.126 & 0.957 & 5.081 & 0.203 & 0.853 & \bf{0.953} & 0.979 \\       
    & 0.5 & 32 & 8 & 0.126 & 0.933 & 5.039 & 0.203 & 0.853 & 0.952 & 0.979 \\
    & 0.5 & 16 & 8 & 0.125 & 0.933 & 5.039 & 0.203 & 0.853 & 0.952 & 0.979 \\
    & 0.5 & 32 & 4 & 0.125 & 0.938 & 5.006 & 0.202 & 0.854 & \bf{0.953} & \bf{0.980} \\      
    & 0.5 & 16 & 4 & \bf{0.123} & \bf{0.920} & \bf{4.990} & \bf{0.201} & \bf{0.858} & \bf{0.953} & \bf{0.980} \\ \hline
  \end{tabular}
  }
\end{center}
\end{table*}
\subsection{Additional Visualization of Estimated Depth Image}
\begin{figure*}[t]
  \begin{center}
      \includegraphics[width=0.99\hsize]{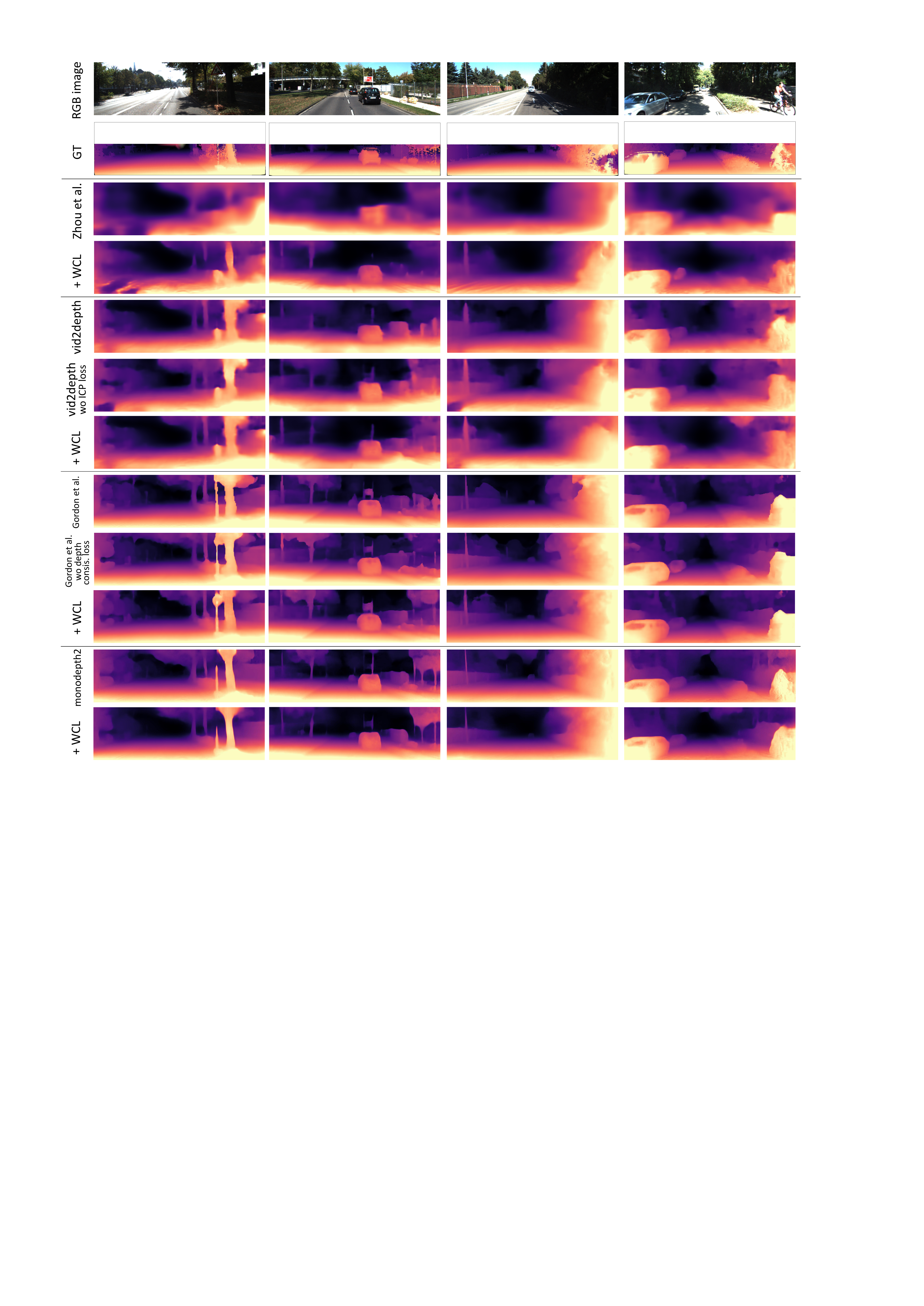}
  \end{center}
	\caption{\small {\bf Additional qualitative results of our proposed method (1).} The top row displays RGB images, the second row displays interpolated ground truth depth images, and the other rows display the estimated depth images, with and without our WCL. As the ground truth is sparse, we interpolate only for visualization. All estimated images are estimated from 416$\times$128 RGB images.}
  \label{f:pred_depth_1}
\end{figure*}

\begin{figure*}[t]
  \begin{center}
      \includegraphics[width=0.99\hsize]{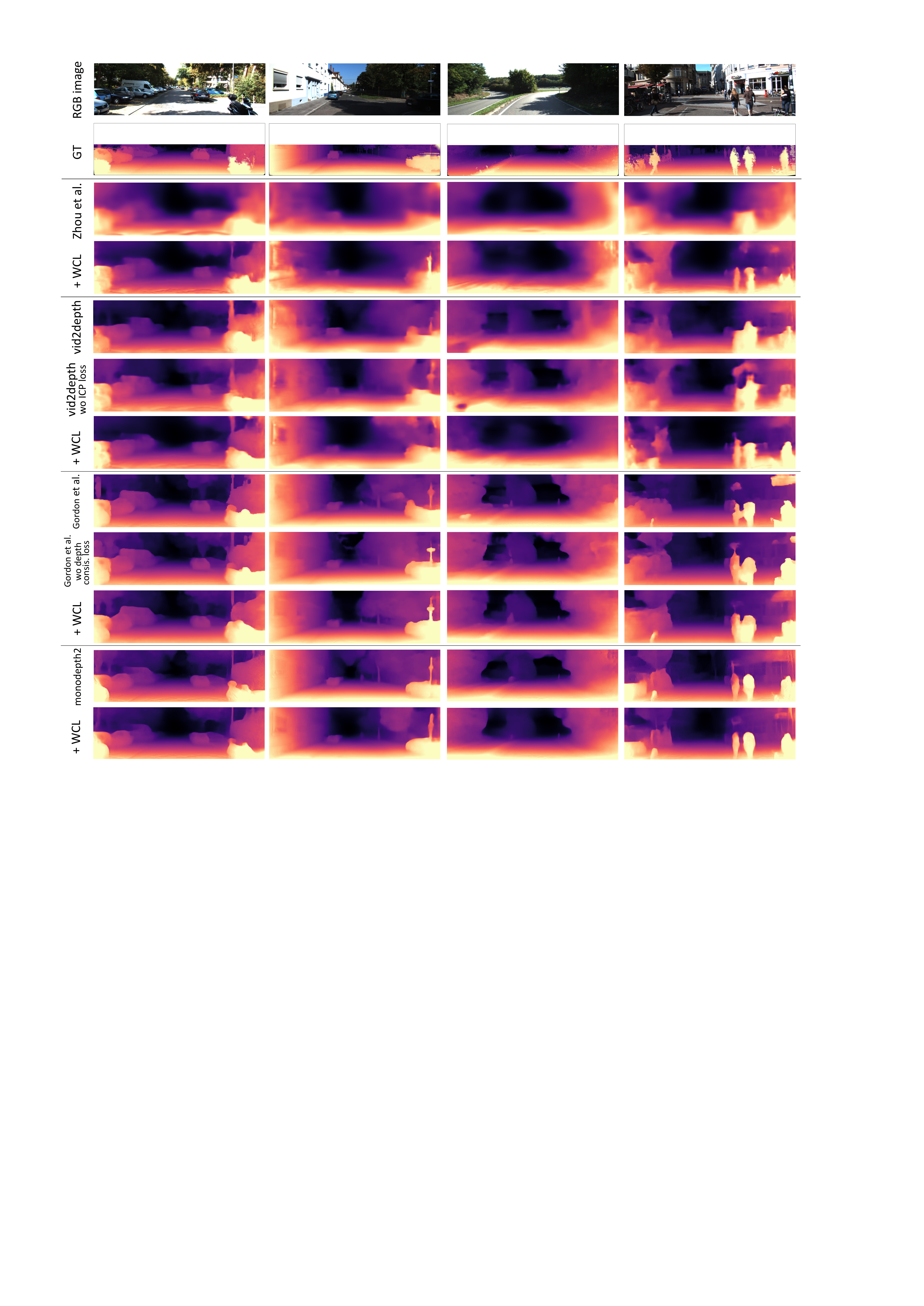}
  \end{center}
	\caption{\small {\bf Additional Qualitative results of our proposed method (2).} The top row displays RGB images, the second row displays interpolated ground truth depth images, and the other rows display the estimated depth images, with and without our WCL. As the ground truth is sparse, we interpolate only for visualization.  All estimated images are estimated from 416$\times$128 RGB images.}
  \label{f:pred_depth_2}
\end{figure*}

To clarify the benefits of our WCL visually, we display the depth images for eight additional cases in Fig.~\ref{f:pred_depth_1} and \ref{f:pred_depth_2}.
Fig.~\ref{f:pred_depth_1} and \ref{f:pred_depth_2} show interpolated ground truth(GT) and estimated depth images by various methods, including four original works by \cite{zhou2017unsupervised,mahjourian2018unsupervised,gordon2019depth,godard2019digging}, and these prior works with our WCL.
Our WCL can sharpen the estimated depth image against the original methods and reduce any artifacts.
